\documentclass[conference]{IEEEtran}

\usepackage{array}
\usepackage{threeparttable}
\usepackage{graphicx}
\usepackage{amsmath}

\newcommand{\bfth}{\mbox{\boldmath{$\theta$}}}

\usepackage{stfloats} 

\usepackage{graphicx}
\usepackage{caption}
\usepackage{subcaption}
\DeclareMathOperator*{\argmax}{arg\,max}
\usepackage[ruled,vlined]{algorithm2e}

\IEEEoverridecommandlockouts    

\begin{document}

\title{\ \\ \LARGE\bf FSMJ: Feature Selection with Maximum Jensen-Shannon Divergence for Text Categorization\thanks{Bo Tang and Haibo He are with the Department of Electrical, Computer and Biomedical Engineering at the University of Rhode Island, Kingston, RI, USA, 02881. E-mail: \{btang, he\}@ele.uri.edu}}

\author{Bo Tang and Haibo He}


\maketitle

\begin{abstract}
In this paper, we present a new wrapper feature selection approach based on Jensen-Shannon (JS) divergence, termed feature selection with maximum JS-divergence (FSMJ), for text categorization. Unlike most existing feature selection approaches, the proposed FSMJ approach is based on real-valued features which provide more information for discrimination than binary-valued features used in conventional approaches. We show that the FSMJ is a greedy approach and the JS-divergence monotonically increases when more features are selected. We conduct several experiments on real-life data sets, compared with the state-of-the-art feature selection approaches for text categorization. The superior performance of the proposed FSMJ approach demonstrates its effectiveness and further indicates its wide potential applications on data mining. 
\end{abstract}


\section{Introduction}
\PARstart{A}{utomated} text categorization is of great interest in many applications and has drawn attention from many researchers in a wide scientific areas \cite{aggarwal2012mining}\cite{aggarwal2012survey}\cite{johnson2014effective}, such as statistics, machine learning, information retrieval and management, to name a few. Considering the text categorization as the classification problem, many existing classification algorithms in machine learning, such as artificial neural network, support vector machine and Bayesian learning, would be easily applied to address this problem. Among these classification algorithms, naive Bayes has gained remarkable success and popularity in text categorization. 

In text categorization, the document is typically represented with the concept of ``the bag of words": each feature corresponds to a term or a phrase in a dictionary collected for a given data set. Tens of thousands terms or phrases in a document lead to a big challenge of the learning from high dimensional data. The ``curse of dimensionality" not only leads to a high computational burden of learning, but also hurts the performance due to irrelevant and redundant features. To address this issue, many feature selection approaches have been proposed to reduce the feature size, such as mutual information, information gain, Chi-squared statistic, etc. Extensive experimental studies on these feature selection methods have shown that most of these methods are able to reduce the computational cost and speed up the learning process without hurting the learning performance too much \cite{yang1997comparative}\cite{forman2003extensive}. However, it has also been shown that none of these methods could consistently perform better than others. 

In this paper, we develop a new feature selection approach based on the maximum of Jensen-Shannon (JS) divergence, termed FSMJ, for text categorization. Unlike most of previous feature selection approaches with binary-valued features, the proposed FSMJ approach is based on real-valued features which are the number of times that the term or the phrase occurs in the document. The proposed FMSJ is a greedy approach, and we show that the JS-divergence monotonically increases when more features are selected. We evaluate the performance of the proposed approach on real-life data sets when the multinomial Naive Bayes (MNB) is used as classifier, compared with several state-of-the-art feature selection methods. Experimental results show a consistent improvement of our proposed approach.

The rest of the paper is organized as follows: In Section II, we introduce the background and the related work on MNB classifiers and feature selection techniques for automatic text categorization. In Section III, we present the properties of divergence measures between two multinomial distributions and introduce our new feature selection approach. Experimental results are described in Section IV along with the performance analysis compared with the state-of-the-art feature selection methods. A conclusion is given in Section V.

\section{Background and Related Work}
\subsection{Background}
Naive Bayes classifier is widely used for text categorization because of its simplicity and efficiency. It follows a basic assumption that the occurrences of each individual term or phrase in a document are mutually independent. A classification decision is made according to the \textit{maximum a posteriori} (MAP) rule for a new document with unknown topic. Several types of naive Bayes classifier have been well studied in literature, such as Bernoulli naive Bayes model, multinomial naive Bayes model, and Poisson naive Bayes model. Among these naive Bayes models, it has been shown that the multinomial model outperforms others \cite{aggarwal2012survey}\cite{mccallum1998comparison}. Specifically, the multinomial model uses ``the bag of words" to represent a document and captures word frequency in a document. A document is considered as a sequence of words sampled from a large vocabulary according to a multinomial distribution.

Let the variable $c \in \mathcal{C} = \{c_1, c_2, \cdots, c_N \}$ be the class label for a $N$-class classification problem, and let the vector $\mathbf{x} = [x_1, x_2, \cdots, x_M]^T$ be the document with $M$ features, where $x_m$ denotes the occurrence of the $m$-th term in the vocabulary $V$. In other words, the vocabulary $V$ is consisted of $M$ terms, and any document is represented by a $M \times 1$ feature vector. In multinomial model, the vector $\mathbf{x}$ follows a multinomial distribution, which is given by:
\begin{align}
\label{mnb}
p(\mathbf{x} | c = c_i; \bfth_i) = \frac{n_{m}!}{x_1 ! x_2 ! \cdots x_M!} \prod_{m=1}^{M} p_{im}^{x_m} 
\end{align}
where $n_m = \sum_{m=1}^M x_m$ denotes the total number of terms in the document and $p_{im}$ is the cell probability for the $m$-th term in class $c_i$. For each class, we have a $M \times 1$ parameter vector $\bfth_i = [p_{i1}, p_{i2}, \cdots, p_{im}]$, where $0 \leq p_{im} \leq 1$ and $\sum_{m}p_{i,m} = 1$ for $i=1,2,\cdots, N$. 

The model parameters $\bfth_i$ are usually estimated from the given training data as the prior knowledge. Let $\mathcal{D}$ be the training data set with $|\mathcal{D}|$ documents in total, and let $z_{ki}$ be the indicator variable which equals 1 when the $k$-th document in $\mathcal{D}$ has the class label $c_i$. Hence, for a multinomial distribution, the maximum likelihood estimate (MLE) of the cell probability of the $m$-th term $p_{im}$ is given by:
\begin{align}
\hat{p}_{im} = \frac{\sum_{k=1}^{|\mathcal{D}|} x_{km} z_{ki}}{\sum_{m=1}^{M} \sum_{k=1}^{|\mathcal{D}|} x_{km} z_{ki}}
\end{align}
where $x_{km}$ denotes the number of times that the $m$-th term appears in the $k$-th document in $\mathcal{D}$. The numerator in above equation is the total number of time the $j$-term appears among documents in class $c_i$, and the denominator is the total number of terms in class $c_i$. 

For a new document to be classified $\mathbf{x} = [x_1, x_2, \cdots, x_M]$, the multinomial naive Bayes makes a decision according to the MAP rule as follows:
\begin{align}
\label{mnb_map}
c^{*} & = \arg\max_{c_i \in \mathcal{C}} p(c_i|\mathbf{x}) \nonumber \\
& = \arg\max_{c_i \in \mathcal{C}} p(\mathbf{x} | c_i; \bfth_i) p(c_i) \nonumber \\
& = \arg\max_{c_i \in \mathcal{C}} \sum_{m=1}^{M} x_m \log p_{im} + \log p(c_i) 
\end{align}
where the class priors $p(c_i)$ are estimated from the training data with the MLE of $\hat{p}(c_i) = \sum_{k=1}^{|\mathcal{D}|}z_{ki} / |\mathcal{D}|$. 

\subsection{Related Work}

Many feature selection methods have been proposed in general machine learning fields, such as regression, classification, and clustering. Some of those methods can be also used for text categorization which can be considered as a multi-class classification problem. Relevance of features is a major concern for designing feature selection methods \cite{joachims1998text} \cite{molina2002feature}. For example, several well-recognized feature selection methods have been developed considering the entropic relevance, such as document frequency, information gain \cite{mitchell1997machine}, mutual information \cite{wiener1995neural}, $\chi^2$ statistic, ect. In \cite{yang1997comparative}, a comparative analysis of these methods is presented. In \cite{forman2003extensive}, an extensive empirical study is performed using these feature selection methods for text categorization.
The empirical results show that feature selection methods can effectively reduce the computation of learning and speed up the learning process with little loss of discriminative performance \cite{forman2003extensive} \cite{dasgupta2007feature}. To find a suitable feature subset for a learning algorithm, several feature selection methods are usually needed to test and compare. It is difficult to select the optimal feature subset in a theoretical way. 

\section{Proposed Feature Selection Approach}
\subsection{Divergence Measures between Two Multinomial Distributions}
Considering a two-class classification problem, each class is represented by a multinomial distribution, saying $P_1=p(\mathbf{x}|c_1;\bfth_1)$ for class $c_1$ and $P_2 = p(\mathbf{x} | c_2; \bfth_2)$ for $c_2$, specified by the probabilities of two $M$-category populations $\bfth_i = \{p_{i1}, p_{i2}, \cdots, p_{iM}\}$, $i=1,2$. According to the information theory \cite{kullback1997information}, Kullback–Leibler (KL) divergence can be used to measure the information for discriminating two distributions. Specifically, we use the KL-divergence $\mathcal{KL}(P_1:P_2)$ to measure the discriminative information of data drawn from class $c_1$ for $P_1$ against $P_2$, and use the KL-divergence $\mathcal{KL}(P_2:P_1)$ to measure the discriminative information of data drawn from class $c_2$ for $P_2$ against $P_1$. According to the definition of the KL-divergence, we have
\begin{align}
\mathcal{KL}(P_1:P_2) = \int_{\mathbf{x}} p(\mathbf{x}|c_1) \log \frac{p(\mathbf{x}|c_1)}{p(\mathbf{x}|c_2)} d \mathbf{x}
\end{align}
Replacing $p(\mathbf{x}|c_i)$ with Eq. (\ref{mnb}), $KL(P_1:P_2)$ can be written as
\begin{align}
\mathcal{KL}(P_1:P_2) & = \int_{\mathbf{x}} p(\mathbf{x}|c_1) \sum_{m=1}^M x_{m} \log \frac{p_{1m}}{p_{2m}} d\mathbf{x} \nonumber \\
& = \sum_{m=1}^M \log \frac{p_{1m}}{p_{2m}} E_{P_1}[x_{m}] 
\end{align}
where $E_{P_1}[x_{m}]$ is the expectation of $x_m$ with respect to the distribution of $p( \mathbf{x} | c_1)$. Since $E_{P_1}[x_{m}] = N p_{1m}$ for a multinomial distribution, we have
\begin{align}
\label{kl_form1}
\mathcal{KL}(P_1:P_2) = \frac{1}{N} \sum_{m=1}^M p_{1m} \log \frac{p_{1m}}{p_{2m}}
\end{align}
Similarly, the KL-divergence $\mathcal{KL}(P_2:P_1)$ has the form of 
\begin{align}
\label{kl_form2}
\mathcal{KL}(P_2:P_1) = \frac{1}{N}\sum_{m=1}^M p_{2m} \log \frac{p_{2m}}{p_{1m}}
\end{align}

We note here that the KL-divergence between two multinominal distributions can be easily calculated with the computational complexity of $\mathcal{O}(M)$. Unlike the conventional feature selection approaches with binary-valued features, we use the real-valued features which retain more discriminative information for classification. The following statement illustrates the capacity of using the KL-divergence measure for discriminating two multinomial distributions.

\newtheorem{remark}{\bf Remark}
\begin{remark}
\label{remark1}
\it Under the MAP rule, the KL-divergence $\mathcal{KL}(P_1:P_2)$ is the measure of discriminative capability for $P_1$ against $P_2$. Given two distributions $P_1$ and $P_2$ and a threshold for classification, a larger value of $\mathcal{KL}(P_1:P_2)$ asymptotically leads to a lower misclassification error for the data drawn from $P_1$.
\end{remark}
\begin{proof}
According to the MAP rule in Eq. (\ref{mnb_map}), any observation $\mathbf{x}_1 = [x_{11}, x_{12}, \cdots, x_{1M}]$ drawn from class $c_1$ is correctly assigned to class $c_1$ if and only if 
\begin{align} 
\sum_{m=1}^{M} x_{1m} \log p_{1m} + \log p(c_1) > \sum_{m=1}^{M} x_{1m} \log p_{2m} + \log p(c_2)
\end{align}
or
\begin{align}
\sum_{m=1}^{M} x_{1m} \log p_{1m} - \sum_{m=1}^{M} x_{1m} \log p_{2m} >  \log p(c_2) - \log p(c_1)
\end{align}
Let $\gamma = \log p(c_2) - \log p(c_1)$ be a threshold, we have 
\begin{align}
\label{kl_map}
\sum_{m=1}^{M} x_{1m} \log \frac{p_{1m}}{p_{2m}} >  \gamma
\end{align}
Hence, when there are infinite number of data (i.e., asymptotically), we take the expectation in above formula with respect to the distribution of $p(x | c_1)$, and we have 
\begin{align}
\label{remark1_KL}
\mathcal{KL}(P_1:P_2) >  \frac{\gamma}{N} = \gamma^{'}
\end{align}
where $\gamma^{'} = \gamma / N$. It can be seen that, for the given two distributions $P_1$ and $P_2$ and a threshold $\gamma^{'}$, a larger $\mathcal{KL}(P_1:P_2)$ indicates less data drawn from the distribution $P_1$ is misclassified in an asymptotic way.
%
\end{proof}

\begin{remark}
\label{remark_2}
\it Under the MAP rule, the Jeffreys divergence (J-divergence) \cite{jeffreys1946invariant}, which is defined by
\begin{align}
\label{JK}
\mathcal{J}(P_1,P_2) = \mathcal{KL}(P_1:P_2) + \mathcal{KL}(P_2:P_1)  
\end{align}
is a measure of difficulty and capacity for discriminating two multinomial distributions. 
\end{remark}
\begin{proof}
Following the Remark \ref{remark1}, the KL-divergence $\mathcal{KL}(P_2:P_1)$ is also the measure of discriminative capability for $P_2$ against $P_1$, and any observation drawn from class $c_2$ is correctly classified if and only if 
\begin{align}
\mathcal{KL}(P_2:P_1) > - \gamma^{'}
\end{align}
Combining it into Eq. (\ref{remark1_KL}), we have \cite{kullback1997information}
\begin{align}
\mathcal{KL}(P_1:P_2) >  \gamma^{'} > -\mathcal{KL}(P_2:P_1)
\end{align}
Meanwhile, since
\begin{align}
\mathcal{KL}(P_1:P_2) \geq 0, \quad  \mathcal{KL}(P_2:P_1) \geq 0
\end{align}
where the equality is satisfied, if and only if $p_{1m} = p_{2m}$, $i = 1, 2, \cdots, M$, the difference between $\mathcal{KL}(P_1:P_2)$ and $-\mathcal{KL}(P_2:P_1)$ or the sum of $\mathcal{KL}(P_1:P_2) + \mathcal{KL}(P_2:P_1)$, i.e., the J-divergence, measures the difficulty and capacity of discriminating these two multinomial distributions, when using the MAP rule in Eq. (\ref{mnb_map}). 
\end{proof}


\subsection{Jensen-Shannon Divergence}

The purpose of feature selection methods is to determine the most informative features which lead to the best prediction performance. Remark \ref{remark1} and \ref{remark_2} illustrate that both KL-divergence and J-divergence correspond to the recognition performance indicator for a two-class classification problem. 
However, both of them are only defined for two probability distributions. Jensen-Shannon (JS) divergence \cite{lin1991divergence} is the one that can be used to measure multi-distribution divergence, in which the divergences of each individual distribution with a reference distribution are calculated and summed together, defined as follows:

\newtheorem{definition}{\bf Definition}
\begin{definition}
\it Let $\mathcal{P}=\{P_1, P_2, \cdots, P_N\}$ be the set of $N$ distributions. The JS-divergence, denoted by $\mathcal{JS}(P_1, P_2, \cdots, P_N)$, is defined by
\begin{align}
\label{JM}
\mathcal{JS}(P_1, P_2, \cdots, P_N) = \sum_{i=1}^N \mathcal{KL}(P_i: P_0)
\end{align}
where $P_0$ is the reference distribution which is the combination of all $N$ distributions: $P_0 = \sum_{k = 1}^N \pi_{k} P_k$, and $\pi_{k}$ are the class priors.
\end{definition}

The JS-divergence is the sum of $N$ KL-divergences. 
Similar to the J-divergence, the JS-divergence holds many nice properties. For example, the JS-divergence is almost positive definite, i.e., $\mathcal{JS}(P_1, P_2, \cdots, P_N) \geq 0$, with equality if and only if $p_{1m} = p_{2m} = \cdots = p_{Nm}$, $m = 1,2,\cdots,M$. It also holds symmetric property, that is, $\mathcal{JS}(\cdots, P_i, \cdots, P_k, \cdots) = \mathcal{JS}(\cdots, P_k, \cdots, P_i, \cdots)$.

\subsection{Proposed Wrapper Feature Selection Approach} 
Our feature selection approach seeks to select features towards maximum JS-divergence, which can be formulated as a subset selection problem: given a set of $M$ features $\mathcal{F}$, $|\mathcal{F}| = M$, we aim to find a subset $\mathcal{F}^{*} \subset \mathcal{F}$, such that,
\begin{align}
\mathcal{F}^{*} = \arg\max_{\mathcal{S} \subset \mathcal{F}} \mathcal{JS}(P_1, P_2, \cdots, P_N | \mathcal{S})
\end{align}
The optimal solution of this problem is also known as NP-hard, and it is intractable particularly for high dimensional data. 

Motivated by the success of our recent greedy approaches for feature selection \cite{kay2015pdf}\cite{tang2015toward}\cite{tang2016tkde}\cite{bo2016letters}, we here propose a wrapper feature subset selection approach which is termed \textit{feature selection with maximum JS-divergence} (FSMJ) to greedily find the most discriminative features for multi-class classification. This approach will be first to determine which feature of the $M$ features when used to construct a two-feature classification problem produces the maximum JS-divergence. Then, we fix this feature and repeat the process over the remaining features, and a rank ordering of the $M$ features can be produced. In this way, the first selected feature is the one that yields the maximum discriminative capability. Specifically, considering a $N$-class classification, our FSMJ approach proceeds as follows:
\begin{itemize}
\item[1.] For each feature $x_m$, $m=1,2,\cdots, M$, we group the remaining features as one feature, and denote these two features by $x^1_m$ and $\bar{x}^1_m$, respectively. We use the number in the superscript of feature to indicate the current step. The constructed feature $\bar{x}^1_m$ has the cell probability of $\bar{p}_{im} = 1 - p_{im}$ for class $c_i$, $i=1, 2,\cdots, N$. For each class, we build a two-class classification problem using the ``one-vs-all" strategy, and calculate the KL-divergence $\mathcal{KL}_{(x^1_m, \bar{x}^1_m)}(P_i, P^c_i)$ for the $i$-th class. Summing all these $N$ KL-divergence in Eq. (\ref{JM}), we obtain the JS-divergence for the $m$-th feature, denoted by $\mathcal{JS}_{(x^1_m, \bar{x}^1_m)}(P_1, P_2, \cdots, P_N)$. We choose the feature that maximizes the JS-divergence among $M$ features, that is,
\begin{align}
\label{step1}
k = \argmax_{m=1,2,\cdots, M} \mathcal{JS}_{(x^1_m, \bar{x}^1_m)}(P_1, P_2, \cdots, P_N)
\end{align}
and denote the most discriminative feature as $x_k$ and rename it $s_1$. 

\item[2.] Next we omit the feature $x_{k}$ and search for the next most discriminative feature in all sets of of three features given by $(s_1, x^2_m, \bar{x}^2_m)$, $m=1,2, \cdots, M$, and $m \neq k$. The third feature $\bar{x}^2_m$ is grouped by all the features without $s_1$ and $x^2_m$, which has the probability of $\bar{p}_{im} = 1 - p_{im} - p_{km}$ for class $c_i$. Again, for each class, we build a two-class classification problem using the ``one-vs-all" strategy and calculate the JS-divergence among them. We find the next feature by maximizing the following JS-divergence:
\begin{align}
\label{step2}
l = \argmax_{\substack{ m=1,2,\cdots, M, \\ m \neq k}} \mathcal{JS}_{(s_1, x^2_m, \bar{x}^2_m)}(P_1, P_2, \cdots, P_N)
\end{align}
and denote the second most discriminative feature as $x_l$ produced in this step and rename it as $s_2$.
\item[3.] Repeat the procedure in step 2, and we can find the third feature $s_3$. Continuing in this fashion, we produce a rank ordering over the original $M$ features, denoted by $\{s_1, s_2, \cdots, s_M\}$.
\end{itemize}

At the end, the FSMJ approach outputs a rank ordering of $M$ features $\{s_1, s_2, \cdots, s_M\}$. The first ordered feature has the most discriminative capacity for classification. We summarize the algorithm implementation in Algorithm \ref{algorithm: greedy} for multi-class classification. The following Theorem \ref{theorem_JMH} further demonstrates that the JS-divergence monotonically increases as more steps are taken. In other words, the JS-divergence increases with more features are selected for multi-class multinomial distributions.

\begin{algorithm}

   \textbf{Input:}
   \begin{itemize}
   \item The estimated probabilities of each item $\bfth_i = [p_{i1}, p_{i2}, \cdots, p_{iM}]$, $i=1,2,\cdots,N$;
   \item The estimated class prior probabilities, denoted by $p(c_i)$, $i=1,2,\cdots, N$;
   \end{itemize}
   \textbf{Algorithm:}
   \begin{itemize}
   \item[1.] For each feature $x_m$, construct $N$ multinomial distributions with two features denoted by $x^1_m, \bar{x}^1_m$, and calculate the JS-divergence using Eq. (\ref{JM}). Choose the feature $x_k$ with the maximum JS-divergence in Eq. (\ref{step1}), and rename it as $s_1$;
   \item[2.] For each feature without the consideration of those selected features (e.g., $s_1$), construct $N$ multinomial distributions with three features denoted by $s_1, x^2_m, \bar{x}^2_m$, and find a feature that maximizes the JS-divergence in Eq. (\ref{step2}). Rename it as $s_2$;
   \item[3.] Continue the procedure in step 3, and find the $k$-th feature in the $k$-th step;
   \end{itemize}
   \textbf{Output:}
   \begin{itemize}
	\item  A rank ordering of $M$ features: $\{ s_1, s_2, \cdots, s_M \}$.   
   \end{itemize}
   \caption{FSMJ: Feature Selection with Maximum JS-Divergence\label{algorithm: greedy}}
  \end{algorithm}


\newtheorem{theorem1}{\bf Theorem}
\begin{theorem1}
\label{theorem_JMH}
\it The JS-divergence at the $(k+1)$-th step of the Algorithm 1 is larger than the one at the $k$-th step, that is, 
\begin{align}
& \mathcal{JS}_{s_1, \cdots, s_{k+1}, x^{k+1}_{m}, \bar{x}^{k+1}_m} (P_1, P_2, \cdots, P_N) \geq \nonumber \\
&\mathcal{JS}_{s_1, \cdots, s_{k}, x^k_{m}, \bar{x}^k_m} (P_1, P_2, \cdots, P_N)
\end{align}
\end{theorem1}
\begin{proof}
Following the convexity of the divergence measure \cite{kullback1997information}, given a set $E$, we have
\begin{align}
\int_E f_1(x) \log \frac{f_1(x)}{f_2(x)} dx \geq \int_E f_1(x) dx \log \frac{\int_{E} f_1(x) dx}{\int_Ef_2(x) dx}
\end{align}
Thus, for two multinomial distributions, it gives us
\begin{align}
\label{theorem1_ineq}
\sum_{m \in \mathcal{I}} p_{1m} \log \frac{p_{1m}}{p_{2m}}  \geq \sum_{m\in \mathcal{I}} p_{1m}  \log \frac{\sum_{m\in \mathcal{I}} p_{1m}}{\sum_{m\in \mathcal{I}} p_{2m}}
\end{align}
where $\mathcal{I}$ is a feature index set. The above equality is satisfied if and only if
\begin{align}
\frac{p_{1k}}{p_{2k}} = \frac{\sum_{m \in \mathcal{I}} p_{1m}}{\sum_{m \in \mathcal{I}} p_{2m}}, \quad \forall k \in \mathcal{I}
\end{align}
Since the JS-divergence is the sum of $N$ KL-divergences, we next prove that each individual KL-divergence at the $(k+1)$-th step is larger than the one in the $k$-th step. This proof is straightforward using Eq. (\ref{theorem1_ineq}). Denote $i_k$ as the feature index of the feature $s_k$ selected at the $k$-th step. At the $(k+1)$-th step, the calculation of the KL-divergence involves $(k+2)$ features: $k$ ordered features $\{s_1, s_2, \cdots, s_k \}$, one feature $x^{k+1}_m$ to be examined, and the one $\bar{x}^{k+1}_m$ that groups all the remaining terms. At the end of the $k$-th step, the calculation of the KL-divergence involves the features $\{s_1, s_2, \cdots, s_k \}$ and the one that groups all the remaining terms (i.e., the combination of $x^{k+1}_m$ and $\bar{x}^{k+1}_m$ at the $(k+1)$-th step). Hence, according to Eq. (\ref{kl_form1}) and Eq. (\ref{theorem1_ineq}), one can easily conclude that the KL-divergence increases when one feature is split into multiple ones, which further produces our desired results in Theorem 1.
\end{proof}

\section{Experiments and Result Analysis}

\subsection{Experimental Setting} 
For our experiments we use the benchmark of \textsc{Reuters} that has been widely tested in text categorization for performance evaluation. In the original version of the \textsc{Reuters}, 21,578 documents with 135 various topics have been collected. We use the ModApte version of the \textsc{Reuters}, in which the documents assigned to multiple topics are removed. This version data set contains 8,293 documents with 65 topics. Since some of 65 topics have limited documents, we extract two data sets from the \textsc{Reuters}, named \textsc{Reuters-10} and \textsc{Reuters-20}, which consist of the documents of the first 10 and 20 most topics, respectively. In these two data sets, there are 18,933 terms or phrases in the collected dictionary. In our preprocessing stage, we discard those terms or phrases that appear in less than 3 documents, and after that we have the original feature size of 7,789.  

We compare the classification performance of the proposed FSMJ approach with other 6 feature selection approaches that are commonly used in text categorization, including document frequency (DF), information gain (IG), Chi-squared statistic (Chi), relevance score (RS), cross entropy (CET) and NGL coefficient (NGL), when multinomial naive Bayes classifier is employed as the base classifier. Except for the DF \cite{azam2012comparison}, all other 5 approaches measure the binary-valued features and the class to indicate the importance of features. Specifically, we denote the $k$-th binary-valued feature by $x_k \in \{0,1\}$, where $x_k = 0$ means that the term does not appear in the document and $x_k = 1$ means that the term appears in the document, and we denote the $i$-th class by $c_i$. The feature importance measurements in IG, Chi, RS, CET and NGL are defined as follows:

{\small  
  \setlength{\abovedisplayskip}{6pt}
  \setlength{\belowdisplayskip}{\abovedisplayskip}
  \setlength{\abovedisplayshortskip}{0pt}
  \setlength{\belowdisplayshortskip}{3pt}
  \begin{align}
  \label{FS_others}
&\text{IG}(x_k, c_i) = p(x_k, c_i) \log \frac{p(x_k, c_i)}{p(x_k) p(c_i)} + p(\bar{x}_k, c_i) \log \frac{p(\bar{x}_k, c_i)}{p(\bar{x}_k) p(c_i)} \nonumber \\
&\text{Chi}(x_k, c_i) = \frac{\left[ p(x_k, c_i) p(\bar{x}_k, \bar{c}_i) - p(x_k, \bar{c}_i) p(\bar{x}_k, c_i) \right]^2}{p(x_k, c_i) p(x_k, \bar{c}_i) p(\bar{x}_k, c_i) p(\bar{x}_k, \bar{c}_i)} \nonumber \\
&\text{RS}(x_k, c_i) = \log \frac{p(x_k | c_i)}{p(\bar{x}_k|\bar{c}_i)} \nonumber \\
&\text{CET}(x_k, c_i) = p(x_k, c_i) \log \frac{ p(x_k, c_i)}{ p(x_k) p(c_i) } \nonumber \\
&\text{NGL}(x_k, c_i) = \frac{ p(x_k, c_i) p(\bar{x}_k, \bar{c}_i) - p(x_k, \bar{c}_i) p(\bar{x}_k, c_i) }{\sqrt{p(x_k, c_i) p(x_k, \bar{c}_i) p(\bar{x}_k, c_i) p(\bar{x}_k, \bar{c}_i)}} \nonumber 
\end{align}
}

Notice that the above metrics measure a ``local" feature importance for each individual class. To obtain a global measurement for all classes, three global functions, including the sum, the maximum and the weighted average, are commonly used. Mathematically, for a measurement $f(x_k, c_i)$, the following three global functions offer a final score for the $k$-th feature $x_k$: 
\begin{align}
&f_{sum} = \sum_{i=1}^N f(x_k, c_i) \nonumber \\
&f_{max} = \max_{i=1,2,\cdots,N }f(x_k, c_i) \nonumber \\
&f_{avg} = \sum_{i=1}^N w_i f(x_k, c_i) \nonumber 
\end{align}
We compare the performance of these feature selection approaches with all three global functions. In contrast, our proposed FSMJ approach is based on the real-valued features and offers feature ranking order directly without the need of the global operation.

\begin{figure*}[ht]
\captionsetup[subfigure]{labelformat=empty}
  \centering
  \subcaptionbox{(a) \label{r10max}}{\includegraphics[width=5.68cm]{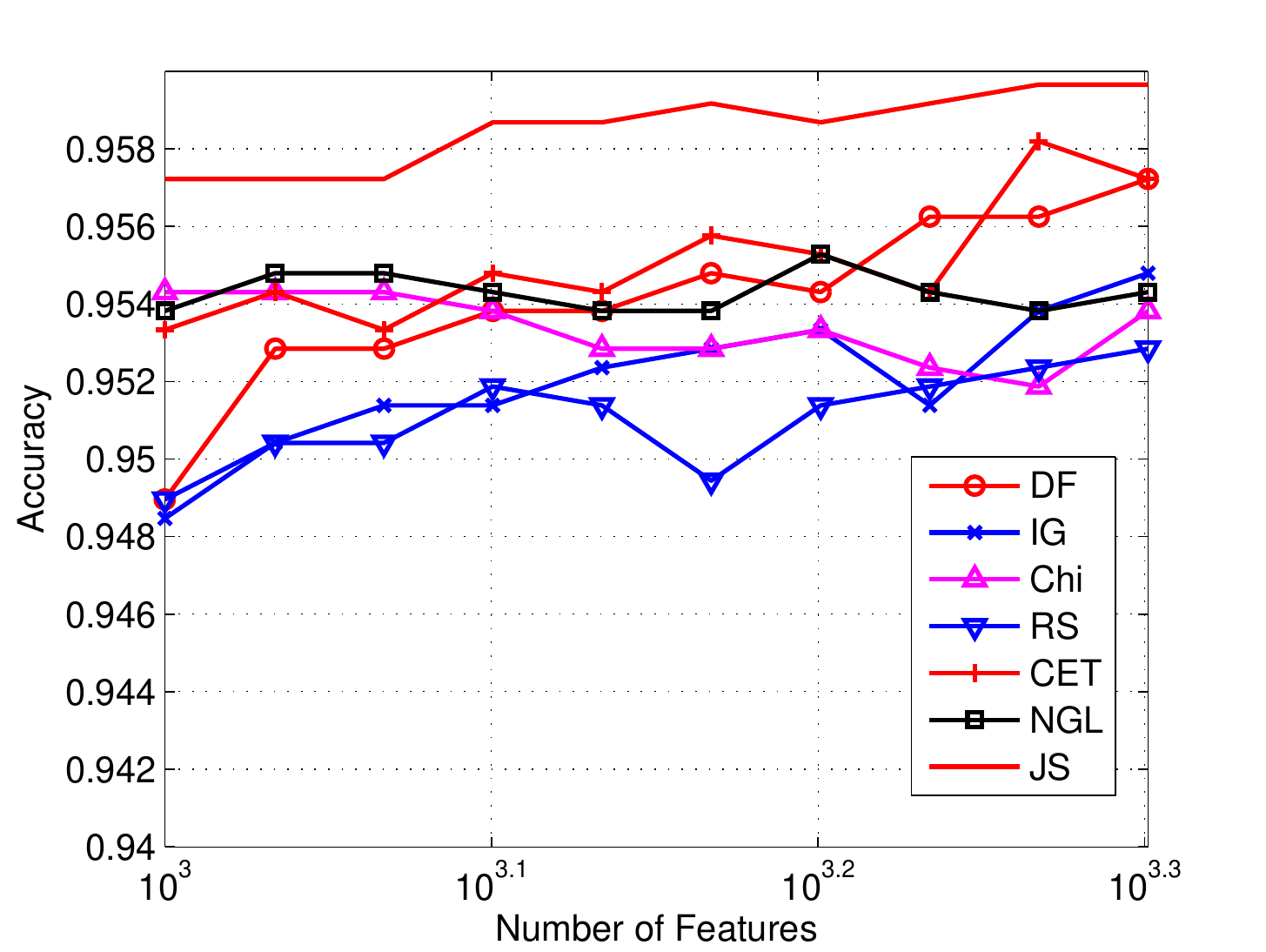}}
  \subcaptionbox{(b) \label{R10sum}}{\includegraphics[width=5.68cm]{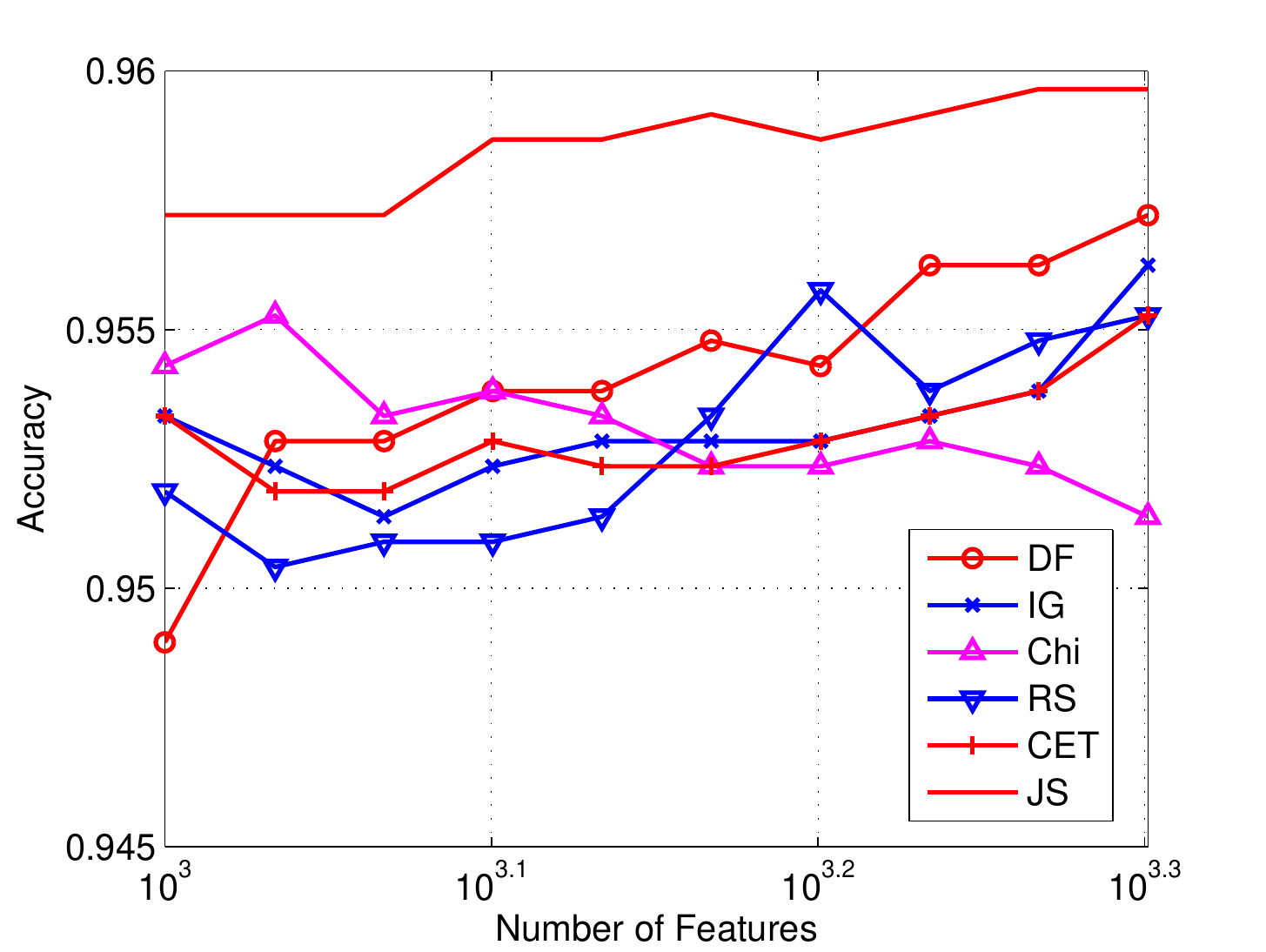}}
  \subcaptionbox{(c) \label{R20avg}}{\includegraphics[width=5.68cm]{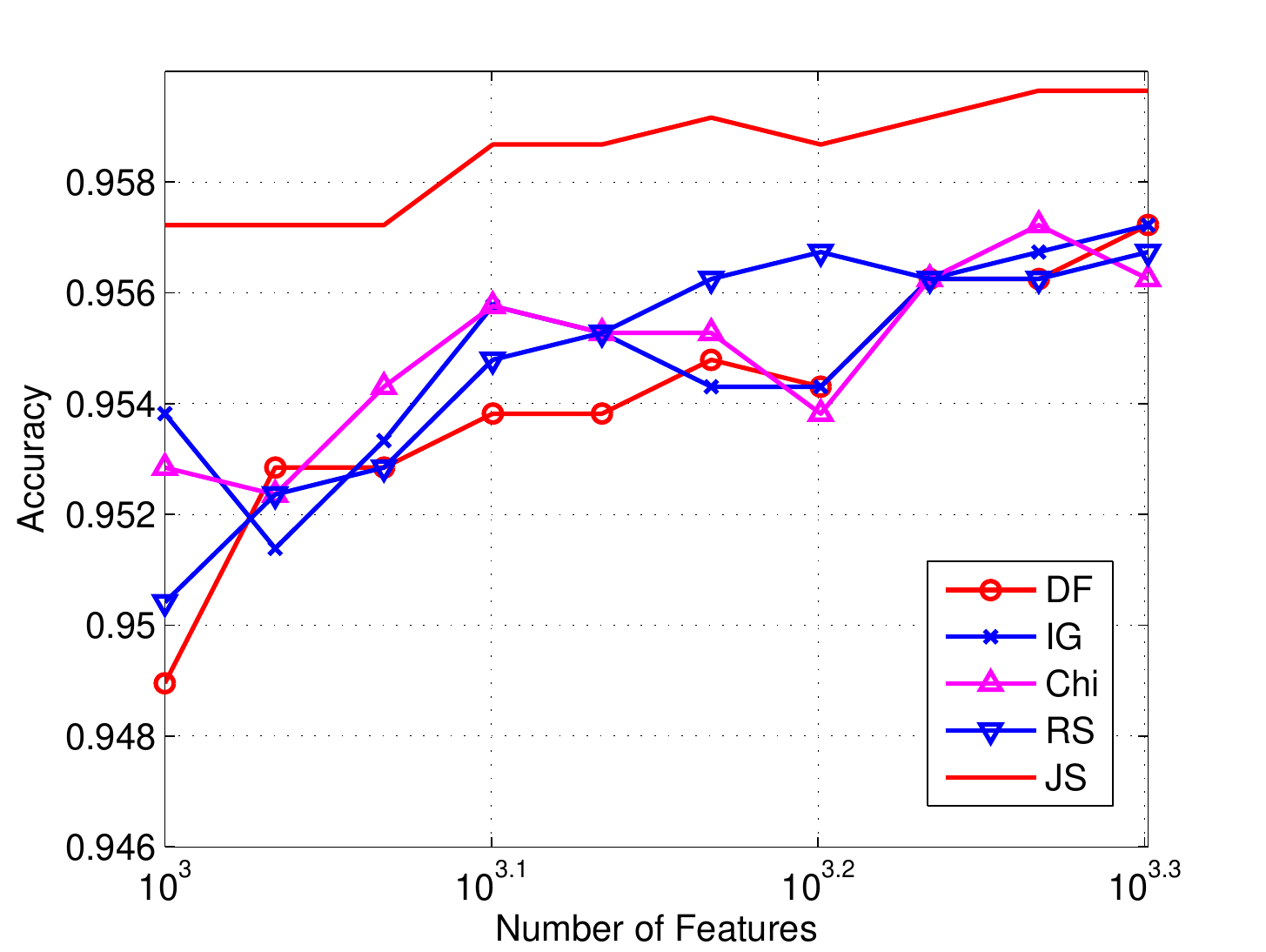}}
   
  \caption{Performance comparison on the \textsc{Reuters-10} data set, when three global operations: (a) the maximum function, (b) the sum function, and (c) the weighted average function, are used for IG, Chi, RS, CET and NGL.}
  \label{all_R1}
\end{figure*}

\begin{figure*}[ht]
\captionsetup[subfigure]{labelformat=empty}
  \centering
  \subcaptionbox{(a) \label{r20max}}{\includegraphics[width=5.68cm]{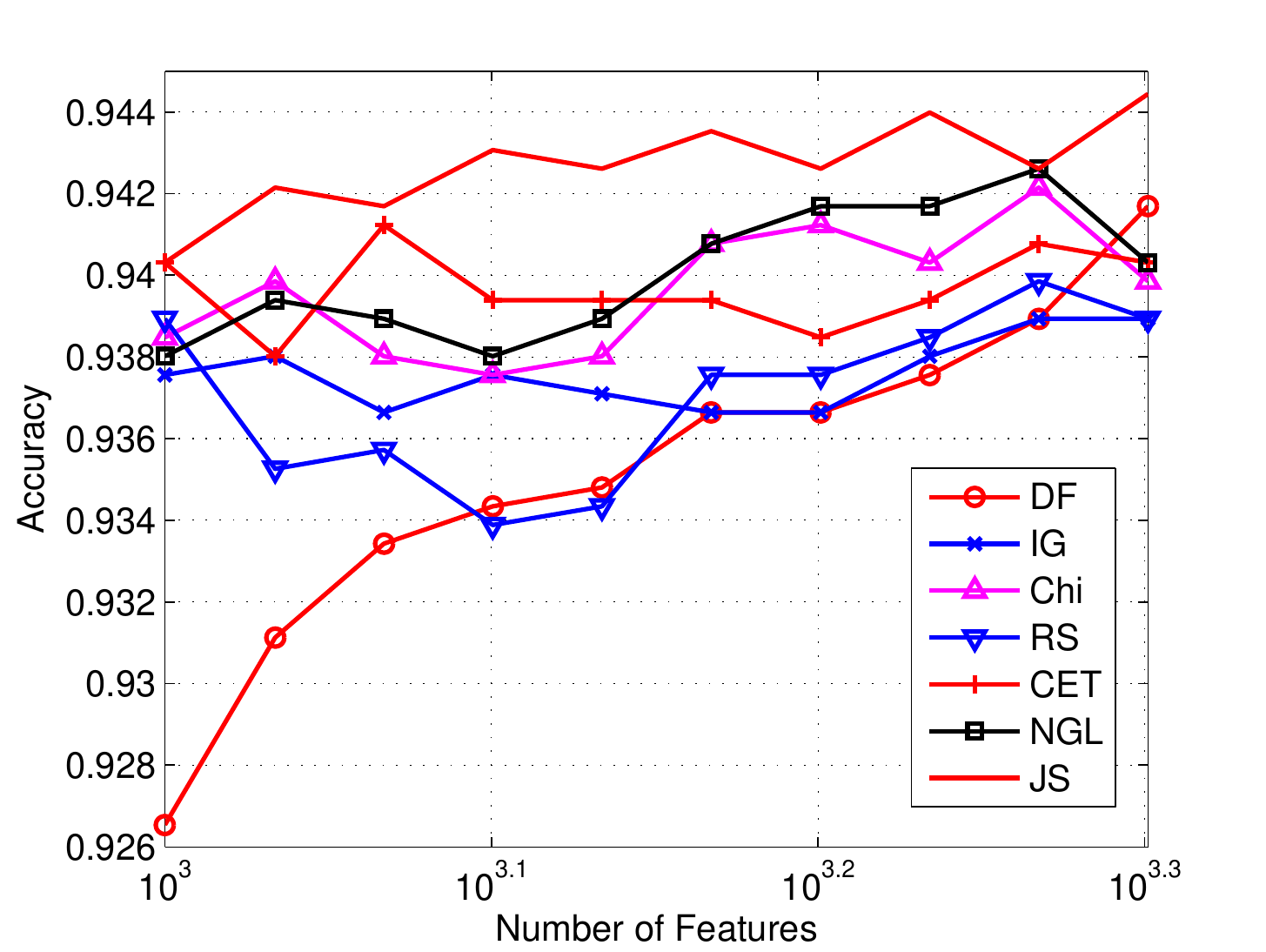}}
  \subcaptionbox{(b) \label{R20sum}}{\includegraphics[width=5.68cm]{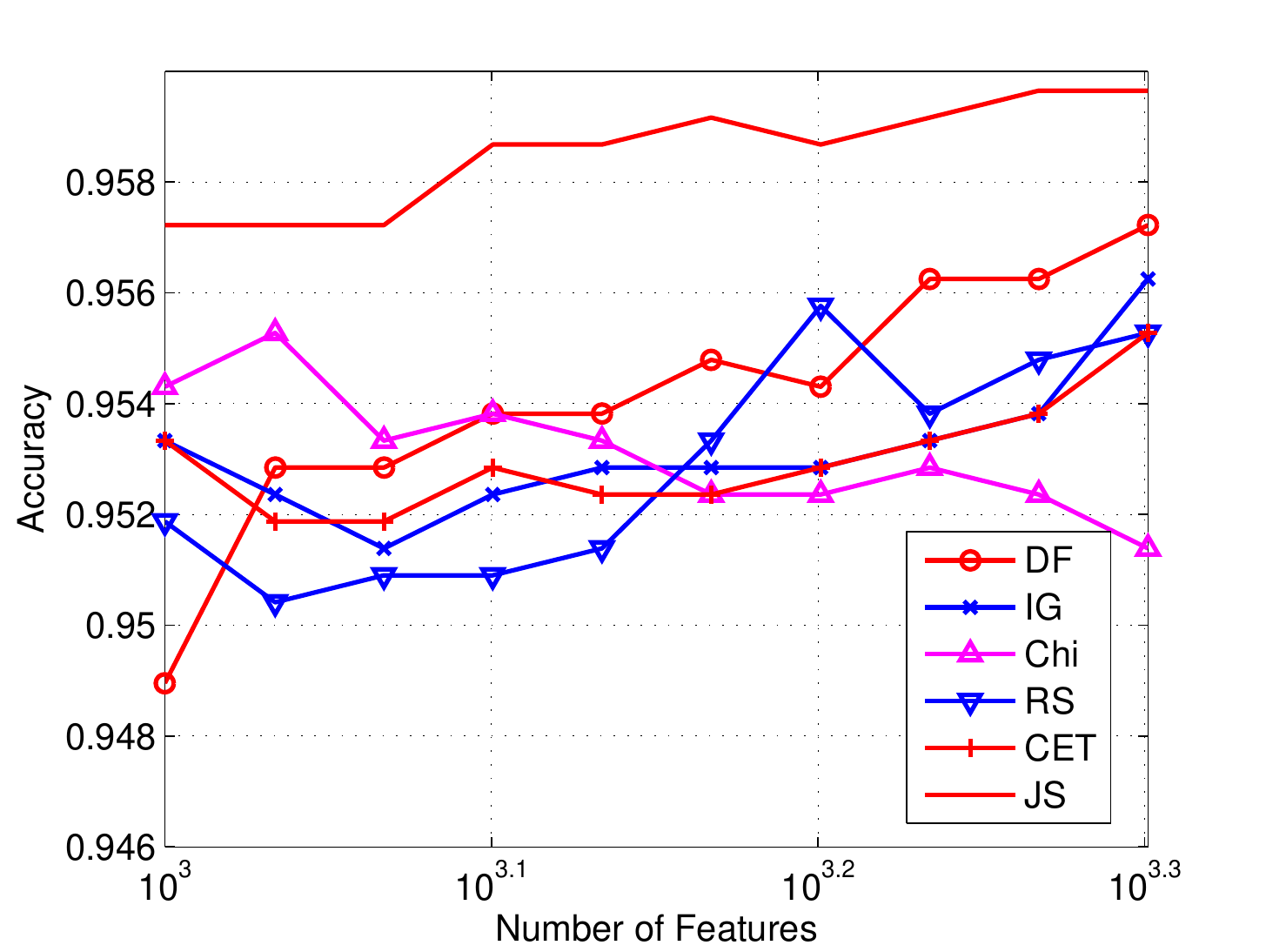}}
  \subcaptionbox{(c) \label{R20avg}}{\includegraphics[width=5.68cm]{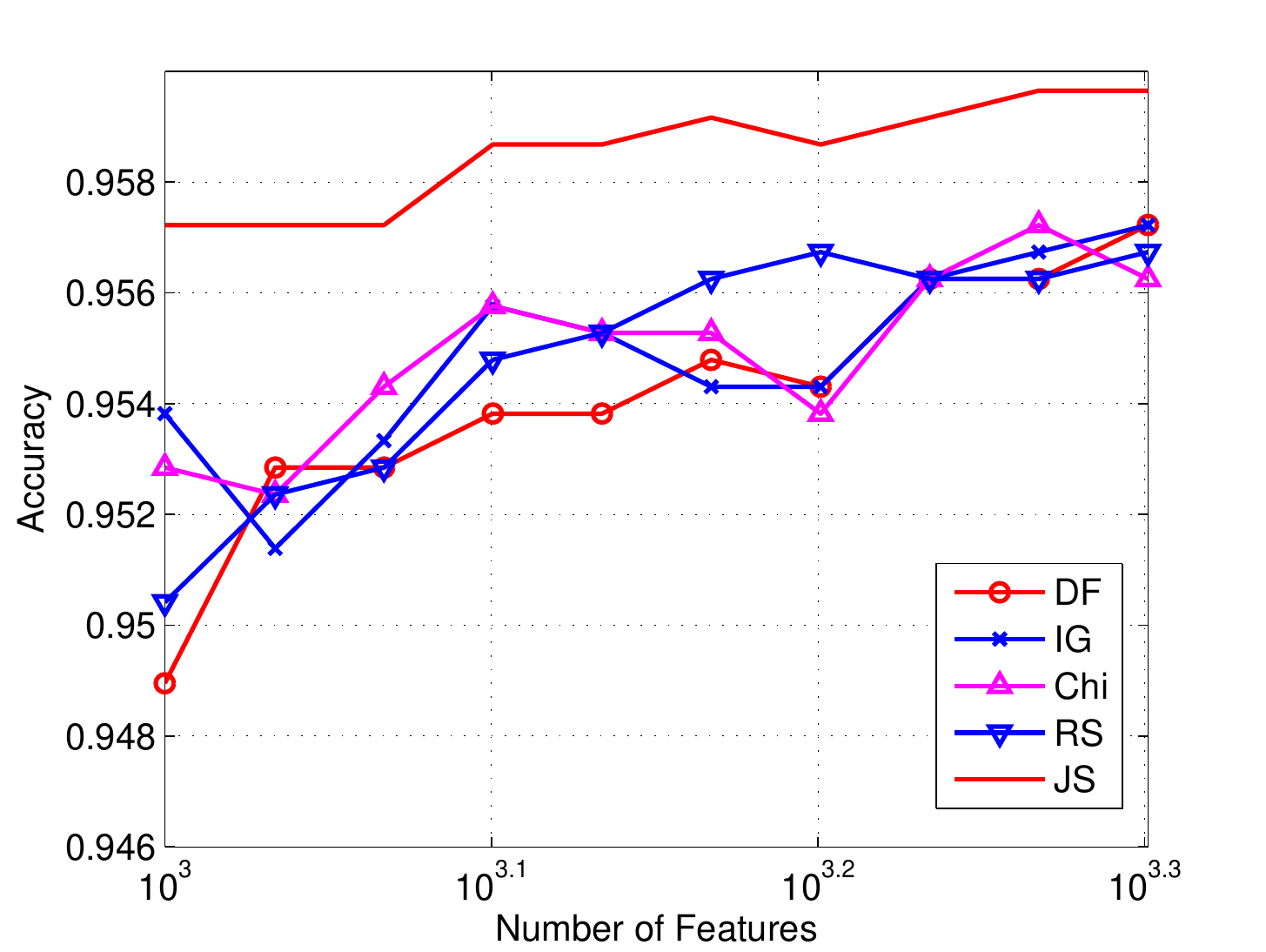}}
   
  \caption{Performance comparison on the \textsc{Reuters-20} data set, when three global operations: (a) the maximum function, (b) the sum function, and (c) the weighted average function, are used for IG, Chi, RS, CET and NGL.}
  \label{all_R2}
\end{figure*}  

\subsection{Result Analysis}
Since a training data set and a testing data set are officially provided in the benchmark of \textsc{Reuters}, we use the given training data set for feature selection and classifier training, and use the given testing data set for performance evaluation. We first show the comparison results on the \textsc{Reuters-10} data set in Fig. \ref{all_R1}, where three global functions are applied for the feature selection approaches of MI, IG, Chi, RS, CET and NGL. It can be shown that our proposed FSMJ approach outperforms all other 6 feature selection approaches with a significant margin. In Fig. \ref{all_R1}(b), we omit the performance of the NGL approach, because the performance of the NGL (the average accuracy is $0.76$) is significantly lower than all other methods. For the same reason, we omit the performance of both CET and NGL in Fig. \ref{all_R1}(c), where the average accuracy of the CET is $0.89$ and the average accuracy of the NGL is $0.77$. 

The experimental results on the \textsc{Reuters-20} data set is shown in Fig. \ref{all_R2}, when the maximum function, the sum function and the weighted average function are used as the global function. It can be shown that the proposed FSMJ approach performs better than all other 6 feature selection approaches with a large margin. For the reason of clear illustration of the comparison, we omit the performance of the NGL in Fig. \ref{all_R2}(b), in which the average accuracy is $0.60$, and we omit the performance of both CET and NGL in Fig. \ref{all_R2}(c), in which the average accuracy of the CET is $0.85$ and the average accuracy of the NGL is $0.73$.

The experimental results in Fig. \ref{all_R1} and \ref{all_R2} show that: (1) the conventional binary-valued feature selection approaches have various performance with different global functions for different data sets. Specifically, the NGL approach with the maximum function usually outperforms others and the NGL approaches with the sum and weighted average functions have poor performance, which is consistent with previous empirical studies  \cite{sebastiani2002machine}; (2) our proposed FSMJ approach has superior performance, compared with all other 6 feature selection approaches, which demonstrates the effectiveness of the proposed approach.

\section{Conclusion}
In this paper, we presented a new feature selection method, termed FSMJ, to rank the order of features based on the maximum Jensen-Shannon divergence. Unlike most of existing methods, the proposed FSMJ approach is based on the real-valued features which retain more discriminative information for measuring feature importance than the binary-valued features. The FSMJ is a greedy approach, and we showed that the JS-divergence monotonically increases when more features are selected for the multinomial distribution. The experimental results demonstrate that our approach has better recognition performance than the state-of-the-art feature selection methods and further indicate wide potential applications on data mining.

\section*{Acknowledgment}
This research was partially supported by National Science Foundation (NSF) under grant ECCS 1053717 and CCF 1439011, and the Army Research Office under grant W911NF-12-1-0378.

\bibliographystyle{ieeetr}
\bibliography{ref}

\end{document}